\documentclass{llncs}

\usepackage[linesnumbered,lined,boxed,vlined]{algorithm2e}
\usepackage{amsfonts}
\usepackage{amsmath}
\usepackage{arydshln}  
\usepackage[table]{xcolor}  
\usepackage{graphics}  
\usepackage{multirow}  
\def\b{\ensuremath\boldsymbol}
\usepackage{graphicx}

\usepackage{amssymb}

\usepackage{url}
\usepackage{tikz}
\usepackage{lipsum}
\newcommand\copyrighttextt{%
  \footnotesize Accepted (to appear) in International Conference on Image Analysis and Recognition (ICIAR) 2020, Springer.}
\newcommand\copyrightnotice{%
\begin{tikzpicture}[remember picture,overlay]
\node[anchor=south,yshift=10pt] at (current page.south) {\fbox{\parbox{\dimexpr\textwidth-\fboxsep-\fboxrule\relax}{\copyrighttextt}}};
\end{tikzpicture}%
}

\begin{document}

\title{Weighted Fisher Discriminant Analysis \\in the Input and Feature Spaces}

\author{Benyamin Ghojogh\inst{1} \and
Milad Sikaroudi\inst{2} \and
H.R. Tizhoosh\inst{2} \and
Fakhri Karray\inst{1} \and
Mark Crowley\inst{1}
}
\authorrunning{F. Author et al.}
%
\institute{Department of Electrical and Computer Engineering, \\ University of Waterloo, Waterloo, ON, Canada \and
KIMIA Lab, University of Waterloo, Waterloo, ON, Canada \\
\email{\{bghojogh, msikaroudi, tizhoosh, karray, mcrowley\}@uwaterloo.ca} 
}



\maketitle              

\begin{abstract}
Fisher Discriminant Analysis (FDA) is a subspace learning method which minimizes and maximizes the intra- and inter-class scatters of data, respectively. Although, in FDA, all the pairs of classes are treated the same way, some classes are closer than the others. Weighted FDA assigns weights to the pairs of classes to address this shortcoming of FDA. In this paper, we propose a cosine-weighted FDA as well as an automatically weighted FDA in which weights are found automatically. We also propose a weighted FDA in the feature space to establish a weighted kernel FDA for both existing and newly proposed weights. Our experiments on the ORL face recognition dataset show the effectiveness of the proposed weighting schemes.  
\end{abstract}

\copyrightnotice

\section{Introduction}

Fisher Discriminant Analysis (FDA) \cite{friedman2001elements}, first proposed in \cite{fisher1936use}, is a powerful subspace learning method which tries to minimize the intra-class scatter and maximize the inter-class scatter of data for better separation of classes. 
FDA treats all pairs of the classes the same way; however, some classes might be much further from one another compared to  other classes. In other words, the distances of classes are different. Treating closer classes need more attention  because classifiers may more easily confuse them whereas classes far from each other  are generally easier to separate. 
The same problem exists in Kernel FDA (KFDA) \cite{mika1999fisher} and in most of subspace learning methods that are based on generalized eigenvalue problem such as FDA and KFDA \cite{ghojogh2019roweis}; hence, a weighting procedure might be more appropriate. 

In this paper, we propose several weighting procedures for FDA and KFDA. The contributions of this paper are three-fold: (1) proposing Cosine-Weighted FDA (CW-FDA) as a new modification of FDA, (2) proposing Automatically Weighted FDA (AW-FDA) as a new version of FDA in which the weights are set automatically, and (3) proposing Weighted KFDA (W-KFDA) to have weighting procedures in the feature space, where both the existing and the newly proposed weighting methods can be used in the feature space.

The paper is organized as follows: In Section \ref{section_FDA_and_KFDA}, we briefly review the theory of FDA and KFDA. In Section \ref{section_weighted_FDA}, we formulate the weighted FDA, review the existing weighting methods, and then propose CW-FDA and AW-FDA. Section \ref{section_weighted_KFDA} proposes weighted KFDA in the feature space. In addition to using the existing methods for weighted KFDA, two versions of CW-KFDA and also AW-KFDA are proposed. Section \ref{section_experiments} reports the experiments. Finally, Section \ref{section_conclusion} concludes the paper. 

\section{Fisher and Kernel Discriminant Analysis}\label{section_FDA_and_KFDA}

\subsection{Fisher Discriminant Analysis}

Let $\{\b{x}_i^{(r)} \in \mathbb{R}^d\}_{i=1}^{n_r}$ denote the samples of the $r$-th class where $n_r$ is the class's sample size. Suppose $\b{\mu}^{(r)} \in \mathbb{R}^d$, $c$, $n$, and $\b{U} \in \mathbb{R}^{d \times d}$ denote the mean of $r$-th class, the number of classes, the total sample size, and the projection matrix in FDA, respectively. 
Although some methods solve FDA using least squares problem \cite{zhang2010regularized,diaz2019deep}, the regular FDA \cite{fisher1936use} maximizes the Fisher criterion \cite{xu2006analysis}:
\begin{align}\label{equation_optimization_FDA_criterion}
&\underset{\b{U}}{\text{maximize}} ~~~ \frac{\textbf{tr}(\b{U}^\top \b{S}_B\, \b{U})}{\textbf{tr}(\b{U}^\top \b{S}_W\, \b{U})},
\end{align}
where $\textbf{tr}(\cdot)$ is the trace of matrix. 
The Fisher criterion is a generalized Rayleigh-Ritz Quotient \cite{parlett1998symmetric}.
We may recast the problem to \cite{ghojogh2019fisher}:
\begin{equation}\label{equation_optimization_FDA}
\begin{aligned}
& \underset{\b{U}}{\text{maximize}}
& & \textbf{tr}(\b{U}^\top \b{S}_B\, \b{U}), \\
& \text{subject to}
& & \b{U}^\top \b{S}_W\, \b{U} = \b{I},
\end{aligned}
\end{equation}
where the $\b{S}_W \in \mathbb{R}^{d \times d}$ and $\b{S}_B \in \mathbb{R}^{d \times d}$ are the intra- (within) and inter-class (between) scatters, respectively \cite{ghojogh2019fisher}:
\begin{align}
\b{S}_W &:= \sum_{r=1}^c  \sum_{i=1}^{n_r} n_r (\b{x}_i^{(r)} - \b{\mu}^{(r)}) (\b{x}_i^{(r)} - \b{\mu}^{(r)})^\top = \sum_{r=1}^c n_r\, \breve{\b{X}}_r\, \breve{\b{X}}_r^\top, \\
\b{S}_B &:= \sum_{r=1}^c \sum_{\ell=1}^c n_r\, n_\ell (\b{\mu}^{(r)} - \b{\mu}^{(\ell)}) (\b{\mu}^{(r)} - \b{\mu}^{(\ell)})^\top = \sum_{r=1}^c n_k\, \b{M}_r\, \b{N}\, \b{M}_r^\top,
\end{align}
where $\mathbb{R}^{d \times n_r} \ni \breve{\b{X}}_r := [\b{x}_1^{(r)} - \b{\mu}^{(r)}, \dots, \b{x}_{n_r}^{(r)} - \b{\mu}^{(r)}]$, $\mathbb{R}^{d \times c} \ni \b{M}_r := [\b{\mu}^{(r)} - \b{\mu}^{(1)}, \dots, \b{\mu}^{(r)} - \b{\mu}^{(c)}]$, and $\mathbb{R}^{c \times c} \ni \b{N} := \textbf{diag}([n_1, \dots, n_c]^\top)$. 
The mean of the $r$-th class is $\mathbb{R}^{d} \ni \b{\mu}^{(r)} := (1/n_r) \sum_{i=1}^{n_r} \b{x}_i^{(r)}$.
The Lagrange relaxation \cite{boyd2004convex} of the optimization problem is: $\mathcal{L} = \textbf{tr}(\b{U}^\top \b{S}_B\, \b{U}) - \textbf{tr}\big(\b{\Lambda}^\top (\b{U}^\top \b{S}_W\, \b{U} - \b{I})\big)$, where $\b{\Lambda}$ is a diagonal matrix which includes the Lagrange multipliers. 
Setting the derivative of Lagrangian to zero gives:
\begin{align}
& \frac{\partial \mathcal{L}}{\partial \b{U}} = 2\b{S}_B \b{U} - 2\b{S}_W\b{U} \b{\Lambda} \overset{\text{set}}{=} \b{0} \implies \b{S}_B\, \b{U} = \b{S}_W\, \b{U} \b{\Lambda}, \label{equation_scatter_generalized_eigendecomposition}
\end{align}
which is the generalized eigenvalue problem $(\b{S}_B, \b{S}_W)$ where the columns of $\b{U}$ and the diagonal of $\b{\Lambda}$ are the eigenvectors and eigenvalues, respectively \cite{ghojogh2019eigenvalue}. 
The $p$ leading columns of $\b{U}$ (so to have $\b{U} \in \mathbb{R}^{d \times p}$) are the FDA projection directions where $p$ is the dimensionality of the subspace. Note that $p  \leq \min(d, n-1, c-1)$ because of the ranks of the inter- and intra-class scatter matrices \cite{ghojogh2019fisher}.  

\subsection{Kernel Fisher Discriminant Analysis}

Let the scalar and matrix kernels be denoted by $k(\b{x}_i, \b{x}_j) := \b{\phi}(\b{x}_i)^\top \b{\phi}(\b{x}_j)$ and $\b{K}(\b{X}_1, \b{X}_2) := \b{\Phi}(\b{X}_1)^\top \b{\Phi}(\b{X}_2)$, respectively, where $\b{\phi}(.)$ and $\b{\Phi}(.)$ are the pulling functions.
According to the representation theory \cite{alperin1993local}, any solution must lie in the span of all the training vectors, hence, $\b{\Phi}(\b{U}) = \b{\Phi}(\b{X})\, \b{Y}$ where $\b{Y} \in \mathbb{R}^{n \times d}$ contains the coefficients. 
The optimization of kernel FDA is \cite{mika1999fisher,ghojogh2019fisher}:
\begin{equation}\label{equation_optimization_kernel_FDA}
\begin{aligned}
& \underset{\b{Y}}{\text{maximize}}
& & \textbf{tr}(\b{Y}^\top \b{\Delta}_B\, \b{Y}), \\
& \text{subject to}
& & \b{Y}^\top \b{\Delta}_W\, \b{Y} = \b{I},
\end{aligned}
\end{equation}
where $\b{\Delta}_W \in \mathbb{R}^{n \times n}$ and $\b{\Delta}_B \in \mathbb{R}^{n \times n}$ are the intra- and inter-class scatters in the feature space, respectively \cite{mika1999fisher,ghojogh2019fisher}:
\begin{align}
\b{\Delta}_W &:= \sum_{r=1}^c n_r\, \b{K}_r\, \b{H}_r\, \b{K}_r^\top, \\
\b{\Delta}_B &:= \sum_{r=1}^c \sum_{\ell=1}^c n_r\, n_\ell (\b{\xi}^{(r)} - \b{\xi}^{(\ell)}) (\b{\xi}^{(r)} - \b{\xi}^{(\ell)})^\top = \sum_{r=1}^c n_r\, \b{\Xi}_r\, \b{N}\, \b{\Xi}_r^\top,
\end{align}
where $\mathbb{R}^{n_r \times n_r} \ni \b{H}_r := \b{I} - (1/n_r) \b{1}\b{1}^\top$ is the centering matrix, the $(i,j)$-th entry of $\b{K}_r \in \mathbb{R}^{n \times n_r}$ is $\b{K}_r(i,j) := k(\b{x}_i, \b{x}_j^{(r)})$, the $i$-th entry of $\b{\xi}^{(r)} \in \mathbb{R}^n$ is $\b{\xi}^{(r)}(i) := (1/n_r) \sum_{j=1}^{n_r} k(\b{x}_i, \b{x}_j^{(r)})$, and $\mathbb{R}^{n \times c} \ni \b{\Xi}_r := [\b{\xi}^{(r)} - \b{\xi}^{(1)}, \dots, \b{\xi}^{(r)} - \b{\xi}^{(c)}]$.

The $p$ leading columns of $\b{Y}$ (so to have $\b{Y} \in \mathbb{R}^{n \times p}$) are the KFDA projection directions which span the subspace.
Note that $p  \leq \min(n, c-1)$ because of the ranks of the inter- and intra-class scatter matrices in the feature space \cite{ghojogh2019fisher}.  

\section{Weighted Fisher Discriminant Analysis}\label{section_weighted_FDA}

The optimization of Weighted FDA (W-FDA) is as follows:
\begin{equation}\label{equation_optimization_FDA_weighted}
\begin{aligned}
& \underset{\b{U}}{\text{maximize}}
& & \textbf{tr}(\b{U}^\top \widehat{\b{S}}_B\, \b{U}), \\
& \text{subject to}
& & \b{U}^\top \b{S}_W\, \b{U} = \b{I},
\end{aligned}
\end{equation}
where the weighted inter-class scatter, $\widehat{\b{S}}_B \in \mathbb{R}^{d \times d}$, is defined as:
\begin{align}\label{equation_weighted_between_scatter}
\widehat{\b{S}}_B := \sum_{r=1}^c \sum_{\ell=1}^c \alpha_{r\ell}\, n_r\, n_\ell (\b{\mu}^{(r)} - \b{\mu}^{(\ell)}) (\b{\mu}^{(r)} - \b{\mu}^{(\ell)})^\top = \sum_{r=1}^c n_r\, \b{M}_r\, \b{A}_r\, \b{N}\, \b{M}_r^\top,
\end{align}
where $\mathbb{R} \ni \alpha_{r\ell} \geq 0$ is the weight for the pair of the $r$-th and $\ell$-th classes, $\mathbb{R}^{c \times c} \ni \b{A}_r := \textbf{diag}([\alpha_{r1}, \dots, \alpha_{rc}])$.
In FDA, we have $\alpha_{r\ell}=1,~ \forall r, \ell \in \{1, \dots, c\}$. However, it is better for the weights to be decreasing with the distances of classes to concentrate more on the nearby classes. We denote the distances of the $r$-th and $\ell$-th classes by $d_{r\ell} := ||\b{\mu}^{(r)} - \b{\mu}^{(\ell)}||_2$.
The solution to Eq. (\ref{equation_optimization_FDA_weighted}) is the generalized eigenvalue problem $(\widehat{\b{S}}_B, \b{S}_W)$ and the $p$ leading columns of $\b{U}$ span the subspace.

\subsection{Existing Manual Methods}\label{section_existing_weights}

In the following, we review some of the existing weights for W-FDA.

\hfill \break
\textbf{Approximate Pairwise Accuracy Criterion:}
The Approximate Pairwise Accuracy Criterion (APAC) method \cite{loog2001multiclass} has the weight function:
\begin{align}\label{equation_weight_APAC}
\alpha_{r\ell} := \frac{1}{2\, d_{r\ell}^2} \text{erf}\Big(\frac{d_{r\ell}}{2\sqrt{2}}\Big),
\end{align}
where $\text{erf}(x)$ is the error function:
\begin{align}
[-1, 1] \ni \text{erf}(x) := \frac{2}{\sqrt{\pi}} \int_0^x e^{-t^2} dt. 
\end{align}
This method approximates the Bayes error for class pairs.

\hfill \break
\textbf{Powered Distance Weighting:}
The powered distance (POW) method \cite{lotlikar2000fractional} uses the following weight function:
\begin{align}\label{equation_weight_POW}
\alpha_{r\ell} := \frac{1}{d_{r\ell}^m},
\end{align}
where $m > 0$ is an integer. As $\alpha_{r\ell}$ is supposed to drop faster than the increase of $d_{k\ell}$, we should have $m \geq 3$ (we use $m=3$ in the experiments). 

\hfill \break
\textbf{Confused Distance Maximization:}
The Confused Distance Maximization (CDM) \cite{zhang2012confused} method uses the confusion probability among the classes as the weight function:
\begin{align}\label{equation_weight_CDM}
\alpha_{r\ell} := 
\left\{
    \begin{array}{ll}
        \frac{n_{\ell | r}}{n_r} & \quad \text{if } k \neq \ell, \\
        0 & \quad \text{if } r = \ell,
    \end{array}
\right.
\end{align}
where $n_{\ell|r}$ is the number of points of class $r$ classified as class $\ell$ by a classifier such as quadratic discriminant analysis \cite{zhang2012confused,ghojogh2019linear}. 
One problem of the CDM method is that if the classes are classified perfectly, all weights become zero. Conditioning the performance of a classifier is also another flaw of this method. 

\hfill \break
\textbf{$k$-Nearest Neighbors Weighting:}
The $k$-Nearest Neighbor ($k$NN) method \cite{zhang2013evaluation} tries to put every class away from its $k$-nearest neighbor classes by defining the weight function as
\begin{align}\label{equation_weight_KNN}
\alpha_{r\ell} := 
\left\{
    \begin{array}{ll}
        1 & \quad \text{if } \b{\mu}^{(\ell)} \in \text{$k$NN}(\b{\mu}^{(r)}), \\
        0 & \quad \text{otherwise}.
    \end{array}
\right.
\end{align}
The $k$NN and CDM methods are sparse to make use of the betting on sparsity principle \cite{friedman2001elements,hastie2015statistical}. However, these methods have some shortcomings. For example, if two classes are far from one another in the input space, they are not considered in $k$NN or CDM, but in the obtained subspace, they may fall close to each other, which is not desirable. 
Another flaw of $k$NN method is the assignment of $1$ to all $k$NN pairs, but in the $k$NN, some pairs might be comparably closer. 

\subsection{Cosine Weighted Fisher Discriminant Analysis}

Literature has shown that cosine similarity works very well with the FDA, especially for face recognition \cite{perlibakas2004distance,mohammadzade2012projection}.
Moreover, according to the opposition-based learning \cite{tizhoosh2005opposition}, capturing similarity and dissimilarity of data points can improve the performance of learning. A promising operator for capturing similarity and dissimilarity (opposition) is cosine. Hence, we propose CW-FDA, as a manually weighted method, with cosine to be the weight defined as
\begin{align}\label{equation_cosine_weight}
\alpha_{r\ell} := 0.5 \times \big[1 + \cos\big(\measuredangle (\b{\mu}^{(r)}, \b{\mu}^{(\ell)})\big)\big] = 0.5 \times \big[1 + \frac{\b{\mu}^{(r)\top} \b{\mu}^{(\ell)}}{||\b{\mu}^{(r)}||_2 ||\b{\mu}^{(\ell)}||_2}\big],
\end{align}
to have $\alpha_{r\ell} \in [0, 1]$.
Hence, the $r$-th weight matrix is $\b{A}_r := \textbf{diag}(\alpha_{r\ell}, \forall \ell)$, which is used in Eq. (\ref{equation_weighted_between_scatter}).
Note that as we do not care about $\alpha_{r,r}$, because inter-class scatter for $r=\ell$ is zero, we can set $\alpha_{rr}=0$.

\subsection{Automatically Weighted Fisher Discriminant Analysis}

In AW-FDA, there are $c+1$ matrix optimization variables which are $\b{V}$ and $\b{A}_k \in \mathbb{R}^{c \times c}, \forall k \in \{1, \dots, c\}$ because at the same time where we want to maximize the Fisher criterion, the optimal weights are found. Moreover, to use the betting on sparsity principle \cite{friedman2001elements,hastie2015statistical}, we can make the weight matrix sparse, so we use ``$\ell_0$'' norm for the weights to be sparse. 
The optimization problem is as follows
\begin{equation}\label{equation_optimization_OWFDA}
\begin{aligned}
& \underset{\b{U},\, \b{A}_r}{\text{maximize}}
& & \textbf{tr}(\b{U}^\top \widehat{\b{S}}_B\, \b{U}), \\
& \text{subject to}
& & \b{U}^\top \b{S}_W\, \b{U} = \b{I}, \\
& & & ||\b{A}_r||_0 \leq k, \quad \forall r \in \{1, \dots, c\}.
\end{aligned}
\end{equation}
We use alternating optimization \cite{jain2017non} to solve this problem:
\begin{align}
& \b{U}^{(\tau+1)} := \arg \max_{\b{U}} \Big(\textbf{tr}(\b{U}^\top \widehat{\b{S}}_B^{(\tau)}\, \b{U}) \,\big|\, \b{U}^\top \b{S}_W\, \b{U} = \b{I} \Big), \label{equation_alternating_U} \\
& \b{A}_r^{(\tau+1)} := \arg \min_{\b{A}_r} \Big(\!-\textbf{tr}(\b{U}^{(\tau+1)\top} \widehat{\b{S}}_B\, \b{U}^{(\tau+1)}) \,\big|\, ||\b{A}_r||_0 \leq k \Big), \forall r, \label{equation_alternating_A}
\end{align}
where $\tau$ denotes the iteration. 

Since we use an iterative solution for the optimization, it is better to normalize the weights in the weighted inter-class scatter; otherwise, the weights gradually explode to maximize the objective function. We use $\ell_2$ (or Frobenius) norm for normalization for ease of taking derivatives. 
Hence, for OW-FDA, we slightly modify the weighted inter-class scatter as
\begin{align}
\widehat{\b{S}}_B &:= \sum_{r=1}^c \sum_{\ell=1}^c \frac{\alpha_{r\ell}}{\sum_{\ell'=1}^c \alpha_{r \ell'}^2}\, n_r\, n_\ell (\b{\mu}^{(r)} - \b{\mu}^{(\ell)}) (\b{\mu}^{(r)} - \b{\mu}^{(\ell)})^\top \\
&= \sum_{r=1}^c n_r\, \b{M}_r\, \breve{\b{A}}_r\, \b{N}\, \b{M}_r^\top,
\end{align}
where $\breve{\b{A}}_r := \b{A}_r / ||\b{A}_r||_F^2$ because $\b{A}_k$ is diagonal, and $||.||_F$ is Frobenius norm.

As discussed before, the solution to Eq. (\ref{equation_alternating_U}) is the generalized eigenvalue problem $(\widehat{\b{S}}_B^{(\tau)}, \b{S}_W)$. 
We use a step of gradient descent \cite{nocedal2006numerical} to solve Eq. (\ref{equation_alternating_A}) followed by satisfying the ``$\ell_0$'' norm constraint \cite{jain2017non}. 
The gradient is calculated as follows.
Let $\mathbb{R} \ni f(\b{U}, \b{A}_k) := -\textbf{tr}(\b{U}^{\top} \widehat{\b{S}}_B\, \b{U})$. 
Using the chain rule, we have:
\begin{align}
\mathbb{R}^{c \times c} \ni \frac{\partial f}{\partial \b{A}_r} = \textbf{vec}^{-1}_{c \times c} \Big[(\frac{\partial \breve{\b{A}}_r}{\partial \b{A}_r})^\top (\frac{\partial \widehat{\b{S}}_B}{\partial \breve{\b{A}}_r})^\top \textbf{vec}(\frac{\partial f}{\partial \widehat{\b{S}}_B}) \Big],
\end{align}
where we use the Magnus-Neudecker convention in which matrices are vectorized, $\textbf{vec}(.)$ vectorizes the matrix, and $\textbf{vec}^{-1}_{c \times c}$ is de-vectorization to $c \times c$ matrix. 
We have $\mathbb{R}^{d \times d} \ni \partial f / \partial \widehat{\b{S}}_B = -\b{U}\b{U}^\top$ whose vetorization has dimensionality $d^2$. 
For the second derivative, we have:
\begin{align}
\mathbb{R}^{d^2 \times c^2} \ni \frac{\partial \widehat{\b{S}}_B}{\partial \breve{\b{A}}_r} = n_r\, (\b{M}_r\, \b{N}^\top) \otimes \b{M}_r,
\end{align}
where $\otimes$ denotes the Kronecker product.
The third derivative is:
\begin{align}
\mathbb{R}^{c^2 \times c^2} \ni \frac{\partial \breve{\b{A}}_r}{\partial \b{A}_r} = \frac{1}{||\b{A}_r||_F^2} \Big(\frac{-2}{||\b{A}_r||_F^2}(\b{A}_r \otimes \b{A}_r) + \b{I}_{c^2}\Big).
\end{align}
The learning rate of gradient descent is calculated using line search \cite{nocedal2006numerical}. 

After the gradient descent step, to satisfy the condition $||\b{A}_r||_0 \leq k$, the solution is projected onto the set of this condition. Because $-f$ should be maximized, this projection is to set the $(c-k)$ smallest diagonal entries of $\b{A}_r$ to zero \cite{jain2017non}. In case $k=c$, the projection of the solution is itself, and all the weights are kept.

After solving the optimization, the $p$ leading columns of $\b{U}$ are the OW-FDA projection directions that span the subspace.

\section{Weighted Kernel Fisher Discriminant Analysis}\label{section_weighted_KFDA}

We define the optimization for Weighted Kernel FDA (W-KFDA) as:
\begin{equation}\label{equation_optimization_kernel_FDA_weighted}
\begin{aligned}
& \underset{\b{Y}}{\text{maximize}}
& & \textbf{tr}(\b{Y}^\top \widehat{\b{\Delta}}_B\, \b{Y}), \\
& \text{subject to}
& & \b{Y}^\top \b{\Delta}_W\, \b{Y} = \b{I},
\end{aligned}
\end{equation}
where the weighted inter-class scatter in the feature space, $\widehat{\b{\Delta}}_B \in \mathbb{R}^{n \times n}$, is defined as:
\begin{align}\label{equation_weighted_Delta_B}
\widehat{\b{\Delta}}_B := \sum_{r=1}^c \sum_{\ell=1}^c \alpha_{r\ell}\, n_r\, n_\ell (\b{\xi}^{(r)} - \b{\xi}^{(\ell)}) (\b{\xi}^{(r)} - \b{\xi}^{(\ell)})^\top = \sum_{r=1}^c n_r\, \b{\Xi}_r\, \b{A}_r\, \b{N}\, \b{\Xi}_r^\top.
\end{align}
The solution to Eq. (\ref{equation_optimization_kernel_FDA_weighted}) is the generalized eigenvalue problem $(\widehat{\b{\Delta}}_B, \b{\Delta}_W)$ and the $p$ leading columns of $\b{Y}$ span the subspace.

\subsection{Manually Weighted Methods in the Feature Space}

All the existing weighting methods in the literature for W-FDA can be used as weights in W-KFDA to have W-FDA in the feature space. 
Therefore, Eqs. (\ref{equation_weight_APAC}), (\ref{equation_weight_POW}), (\ref{equation_weight_CDM}), and (\ref{equation_weight_KNN}) can be used as weights in Eq. (\ref{equation_weighted_Delta_B}) to have W-KFDA with APAC, POW, CDM, and $k$NN weights, respectively. 
To the best of our knowledge, W-KFDA is novel and has not appeared in the literature. Note that there is a weighted KFDA in the literature \cite{hamid2012weighted}, but that is for data integration, which is for another purpose and has an entirely different approach. 

The CW-FDA can be used in the feature space to have CW-KFDA. For this, we propose two versions of CW-KFDA: (I) In the first version, we use Eq. (\ref{equation_cosine_weight}) or $\b{A}_r := \textbf{diag}(\alpha_{r\ell}, \forall \ell)$ in the Eq. (\ref{equation_weighted_Delta_B}). 
(II) In the second version, we notice that cosine is based on inner product so the normalized kernel matrix between the means of classes can be used instead to use the similarity/dissimilarity in the feature space rather than in the input space. Let $\mathbb{R}^{d \times c} \ni \b{M} := [\b{\mu}_1, \dots, \b{\mu}_c]$. Let $\widehat{\b{K}}_{i,j} := \b{K}_{i,j} / \sqrt{\b{K}_{i,i} \b{K}_{j,j}}$ be the normalized kernel matrix \cite{ah2010normalized} where $\b{K}_{i,j}$ denotes the $(i,j)$-th element of the kernel matrix $\mathbb{R}^{c \times c} \ni \b{K}(\b{M}, \b{M}) = \b{\Phi}(\b{M})^\top \b{\Phi}(\b{M})$. The weights are $[0,1] \ni \alpha_{r\ell} := \widehat{\b{K}}_{r,\ell}$ or $\b{A}_r := \textbf{diag}(\widehat{\b{K}}_{r,\ell}, \forall \ell)$. We set $\alpha_{r,r}=0$.

\subsection{Automatically Weighted Kernel Fisher Discriminant Analysis}

Similar to before, the optimization in AW-KFDA is:
\begin{equation}\label{equation_optimization_kernel_OWFDA}
\begin{aligned}
& \underset{\b{Y},\, \b{A}_r}{\text{maximize}}
& & \textbf{tr}(\b{Y}^\top \widehat{\b{\Delta}}_B\, \b{Y}), \\
& \text{subject to}
& & \b{Y}^\top \b{\Delta}_W\, \b{Y} = \b{I}, \\
& & & ||\b{A}_r||_0 \leq k, \quad \forall r \in \{1, \dots, c\},
\end{aligned}
\end{equation}
where $\widehat{\b{\Delta}}_B := \sum_{r=1}^c n_r\, \b{\Xi}_r\, \breve{\b{A}}_r\, \b{N}\, \b{\Xi}_r^\top$.
This optimization is solved similar to how Eq. (\ref{equation_optimization_OWFDA}) was solved where we have $\b{Y}\in \mathbb{R}^{n \times d}$ rather than $\b{U} \in \mathbb{R}^{d \times d}$. 
Here, the solution to Eq. (\ref{equation_alternating_U}) is the generalized eigenvalue problem $(\widehat{\b{\Delta}}_B^{(\tau)}, \b{\Delta}_W)$.  
Let $f(\b{Y}, \b{A}_k) := -\textbf{tr}(\b{Y}^{\top} \widehat{\b{\Delta}}_B\, \b{Y})$.
The Eq. (\ref{equation_alternating_A}) is solved similarly but we use $\mathbb{R}^{n \times n} \ni \partial f / \partial \widehat{\b{\Delta}}_B = -\b{Y}\b{Y}^\top$ and
\begin{align}
&\mathbb{R}^{c \times c} \ni \frac{\partial f}{\partial \b{A}_r} = \textbf{vec}^{-1}_{c \times c} \Big[(\frac{\partial \breve{\b{A}}_r}{\partial \b{A}_r})^\top (\frac{\partial \widehat{\b{\Delta}}_B}{\partial \breve{\b{A}}_r})^\top \textbf{vec}(\frac{\partial f}{\partial \widehat{\b{\Delta}}_B}) \Big], \\
&\mathbb{R}^{d^2 \times c^2} \ni \frac{\partial \widehat{\b{\Delta}}_B}{\partial \breve{\b{A}}_r} = n_r\, (\b{\Xi}_r\, \b{N}^\top) \otimes \b{\Xi}_r.
\end{align}
After solving the optimization, the $p$ leading columns of $\b{Y}$ span the OW-KFDA subspace. 
Recall $\b{\Phi}(\b{U}) = \b{\Phi}(\b{X})\, \b{Y}$.
The projection of some data $\b{X}_t \in \mathbb{R}^{d \times n_t}$ is $\mathbb{R}^{p \times n_t} \ni \widetilde{\b{X}}_t = \b{\Phi}(\b{U})^\top \b{\Phi}(\b{X}_t) = \b{Y}^\top \b{\Phi}(\b{X})^\top \b{\Phi}(\b{X}_t) = \b{Y}^\top \b{K}(\b{X}, \b{X}_t)$.

\section{Experiments}\label{section_experiments}

\subsection{Dataset}

\begin{figure}[!h]
\centering
\includegraphics[width=4.8in]{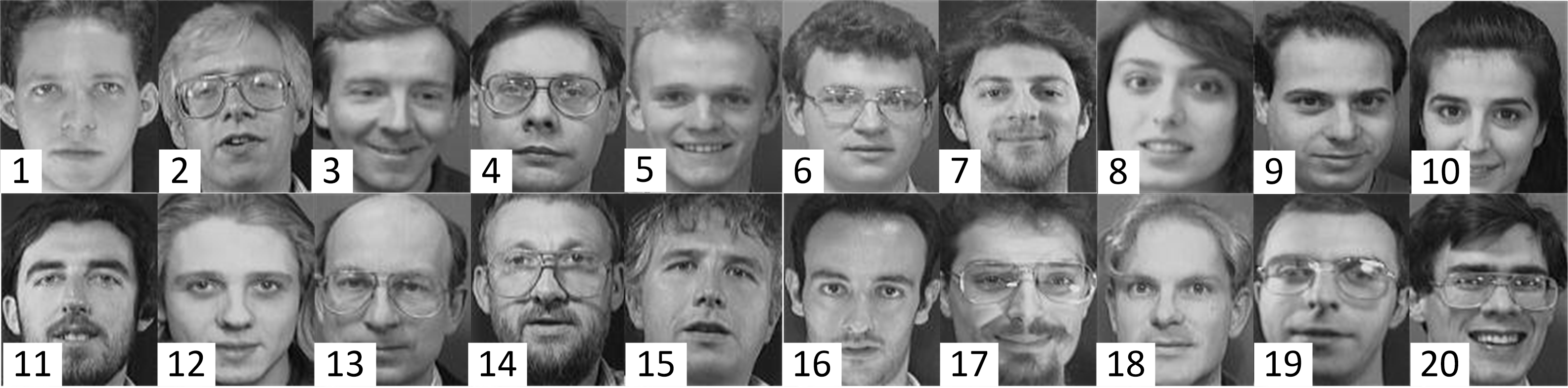}
\caption{Sample images of the classes in the ORL face dataset. Numbers are the class indices.}
\label{figure_dataset}
\end{figure}

For experiments, we used the public ORL face recognition dataset \cite{web_att_dataset} because face recognition has been a challenging task and FDA has numerously been used for face recognition (e.g., see \cite{belhumeur1997eigenfaces,perlibakas2004distance,mohammadzade2012projection}).
This dataset includes 40 classes, each having ten different poses of the facial picture of a subject, resulting in 400 total images. 
For computational reasons, we selected the first 20 classes and resampled the images to $44 \times 36$ pixels. 
Please note that massive datasets are not feasible for the KFDA/FDA because of having a generalized eigenvalue problem in it.
Some samples of this dataset are shown in Fig. \ref{figure_dataset}. 
The data were split into training and test sets with $66\%/33\%$ portions and were standardized to have mean zero and variance one. 
 
\subsection{Evaluation of the Embedding Subspaces}

For the evaluation of the embedded subspaces, we used the 1-Nearest Neighbor (1NN) classifier because it is useful to evaluate the subspace by the closeness of the projected data samples. The training and out-of-sample (test) accuracy of classifications are reported in Table \ref{table_KNN_classification}.
In the input space, $k$NN with $k=1,3$ have the best results but in $k=c-1$, AW-FDA outperforms it in generalization (test) result. The performances of CW-FDA and AW-FDA with $k=1,3$ are promising, although not the best. 
For instance, AW-FDA with $k=1$ outperforms weighted FDA with APAC, POW, and CDM methods in the training embedding, while has the same performance as $k$NN.
In most cases, AW-FDA with all $k$ values has better performance than the FDA, which shows the effectiveness of the obtained weights compared to equal weights in FDA. Also, the sparse $k$ in AWF-FDA outperforming FDA (with dense weights equal to one) validates the betting on sparsity. 

\begin{table*}[!t]
\begin{minipage}{\textwidth}
\caption{Accuracy of $1$NN classification for different obtained subspaces. In each cell of input or feature spaces, the first and second rows correspond to the classification accuracy of training and test data, respectively.}
\label{table_KNN_classification}
\setlength\extrarowheight{5pt}
\centering
\scalebox{0.7}{    
\begin{tabular}{c || c | c | c | c | c | c | c | c | c | c | c | c}
& FDA &  APAC & POW & CDM & $k$NN  & $k$NN  & $k$NN & CW-FDA & CW-FDA & AW-FDA  & AW-FDA  & AW-FDA \\
& &  & &  &  ($k=1$) & ($k=3$) &  ($k=c-1$) & version 1  & version 2 & ($k=1$) & ($k=3$) & ($k=c-1$) \\
\hline
\hline
Input & 97.01\% & 97.01\% & 97.01\% & 74.62\% & 97.76\% & 97.76\% & 97.01\% & 97.01\% & -- & 97.76\% & 97.01\% & 96.26\% \\
space & 92.42\% & 93.93\% & 96.96\% & 45.45\% & 96.96\% & 98.48\% & 92.42\% & 92.42\% & -- & 87.87\% & 93.93\% & 93.93\% \\
\hline
Feature & 97.01\% & 97.01\% & 97.01\% & 91.79\% & 95.52\% & 97.76\% & 97.01\% & 97.01\% & 97.01\% & 100\% & 100\% & 100\% \\
space & 83.33\% & 86.36\% & 89.39\% & 77.27\% & 80.30\% & 83.33\% & 83.33\% & 84.84\% & 87.87\% & 100\% & 100\% & 100\% \\
\hline
\hline
\end{tabular}%
}
\end{minipage}
\end{table*}

\subsection{Comparison of Fisherfaces}

\begin{figure}[!t]
\centering
\includegraphics[width=4.8in]{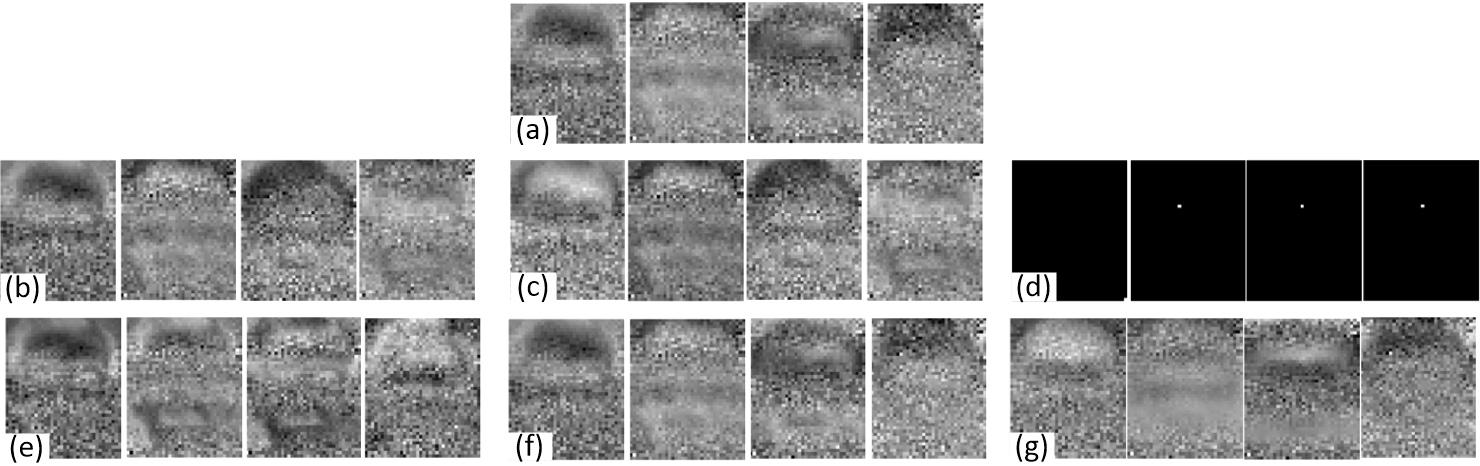}
\caption{The leading Fisherfaces in (a) FDA, (b) APAC, (c) POW, (d) CDM, (e) $k$NN, (f) CW-FDA, and (g) AW-FDA.}
\label{figure_fisherfaces}
\end{figure}

In the feature space, where we used the radial basis kernel, AW-KFDA has the best performance with entirely accurate recognition. 
Both versions of CW-KFDA outperform regular KFDA and KFDA with CDM, and $k$NN (with $k=1, c-1$) weighting. They also have better generalization than APAC, $k$NN with all $k$ values. 
Overall, the results show the effectiveness of the proposed weights in the input and feature spaces. 
Moreover, the existing weighting methods, which were for the input space, have outstanding performance when used in our proposed weighted KFDA (in feature space). This shows the validness of the proposed weighted KFDA even for the existing weighting methods.

\begin{figure}[!t]
\centering
\includegraphics[width=4.8in]{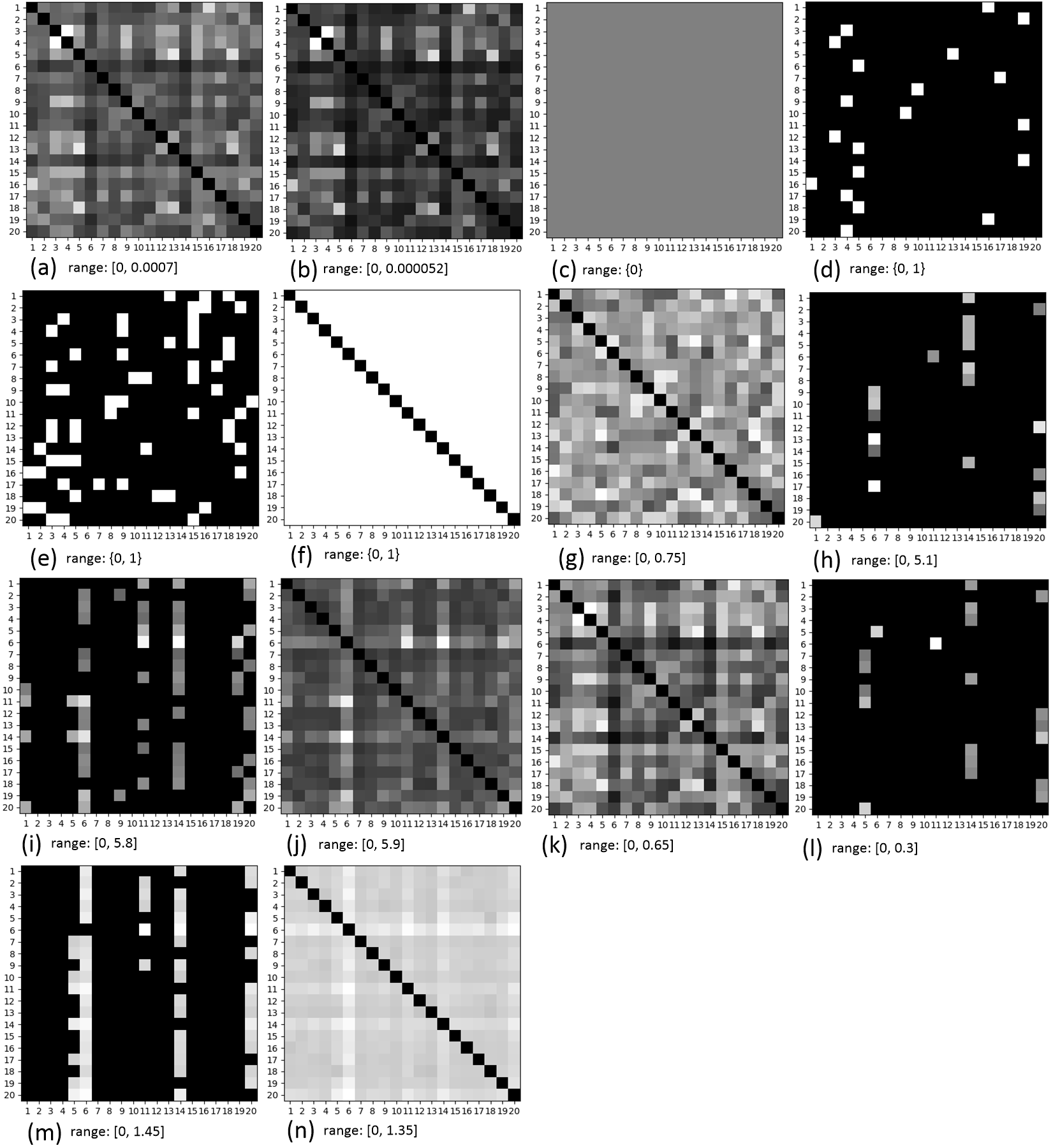}
\caption{The weights in (a) APAC, (b) POW, (c) CDM, (d) $k$NN with $k=1$, (e) $k$NN with $k=3$, (f) $k$NN with $k=c-1$, (g) CW-FDA, (h) AW-FDA with $k=1$, (i) AW-FDA with $k=3$, (j) AW-FDA with $k=c-1$, (k) CW-KFDA, (l) AW-KFDA with $k=1$, (m) AW-KFDA with $k=3$, (n) AW-KFDA with $k=c-1$. The rows and columns index the classes.}
\label{figure_weights}
\end{figure}

Figure \ref{figure_fisherfaces} depicts the four leading eigenvectors obtained from the different methods, including the FDA itself. These ghost faces, or so-called Fisherfaces \cite{belhumeur1997eigenfaces}, capture the critical discriminating facial features to discriminant the classes in subspace. 
Note that Fisherfaces cannot be shown in kernel FDA as its projection directions are $n$ dimensional. 
CDM has captured some pixels as features because its all weights have become zero for its explained flaw (see Section \ref{section_existing_weights} and Fig. \ref{figure_weights}). 
The Fisherfaces, in most of the methods including CW-FDA, capture information of facial organs such as hair, forehead, eyes, chin, and mouth. 
The features of AW-FDA are more akin to the Haar wavelet features, which are useful for facial feature detection \cite{wang2014analysis}. 

\subsection{Comparison of the Weights}

We show the obtained weights in different methods in Fig. \ref{figure_weights}. The weights of APAC and POW are too small, while the range of weights in the other methods is more reasonable. 
The weights of CDM have become all zero because the samples were purely classified (recall the flaw of CDM).
The weights of $k$NN method are only zero and one, which is a flaw of this method because, amongst the neighbors, some classes are closer. This issue does not exist in AW-FDA with different $k$ values. 
Moreover, although not all the obtained weights are visually interpretable, some non-zero weights in AW-FDA or AW-KFDA, with e.g. $k=1$, show the meaningfulness of the obtained weights (noticing Fig. \ref{figure_dataset}). For example, the non-zero pairs $(2, 20)$, $(4, 14)$, $(13, 6)$, $(19, 20)$, $(17, 6)$ in AW-FDA and the pairs $(2, 20)$, $(4, 14)$, $(19, 20)$, $(17, 14)$ in AW-KFDA make sense visually because of having glasses so their classes are close to one another.

\section{Conclusion}\label{section_conclusion}

In this paper, we discussed that FDA and KFDA have a fundamental flaw, and that is treating all pairs of classes in the same way while some classes are closer to each other and should be processed with more care for a better discrimination. We proposed CW-FDA with cosine weights and also AW-FDA in which the weights are found automatically. We also proposed a weighted KFDA to weight FDA in the feature space. We proposed AW-KFDA and two versions of CW-KFDA as well as utilizing the existing weighting methods for weighted KFDA. The experiments in which we evaluated the embedding subspaces, the Fisherfaces, and the weights, showed the effectiveness of the proposed methods.
The proposed weighted FDA methods outperformed regular FDA and many of the existing weighting methods for FDA.
For example, AW-FDA with $k=1$ outperformed weighted FDA with APAC, POW, and CDM methods in the training embedding. In feature space, AW-KFDA obtained perfect discrimination.

\bibliographystyle{splncs}      
\bibliography{references.bib}            

\end{document}